%% file: main.tex
\documentclass[runningheads]{llncs}
\usepackage{graphicx}

\begin{document}
\title{EBOCA: Evidences for BiOmedical Concepts Association Ontology}

\author{Andrea Álvarez Pérez\inst{1}\orcidID{0000-0002-9182-0295} \and
Ana Iglesias-Molina\inst{2}\orcidID{0000-0001-5375-8024} \and
Lucía Prieto Santamaría\inst{1}\orcidID{0000-0003-1545-3515} \and
María Poveda-Villalón\inst{2}\orcidID{0000-0003-3587-0367} \and
Carlos Badenes-Olmedo\inst{2}\orcidID{0000-0002-2753-9917} \and
Alejandro Rodríguez-González\inst{1}\orcidID{0000-0001-8801-4762}}

\authorrunning{A. Álvarez Pérez et al.}

\institute{Centro de Tecnología Biomédica, Universidad Politécnica de Madrid, Madrid, Spain \\
\email{\{andrea.alvarezp\}@alumnos.upm.es}
\email{\{lucia.prieto.santamaria,alejandro.rg\}@upm.es}
\and
Ontology Engineering Group, Universidad Politécnica de Madrid,  Madrid, Spain \\
\email{\{ana.iglesiasm,m.poveda,carlos.badenes\}@upm.es}}
\maketitle              
\begin{abstract}

There is a large number of online documents data sources available nowadays. The lack of structure and the differences between formats are the main difficulties to automatically extract information from them, which also has a negative impact on its use and reuse. 
In the biomedical domain, the DISNET platform emerged to provide researchers with a resource to obtain information in the scope of human disease networks by means of large-scale heterogeneous sources.
Specifically in this domain, it is critical to offer not only the information extracted from different sources, but also the evidence that supports it.

This paper proposes EBOCA, an ontology that describes (i) biomedical domain concepts and associations between them, and (ii) evidences supporting these associations; with the objective of providing an schema to improve the publication and description of evidences and biomedical associations in this domain. The ontology has been successfully evaluated to ensure there are no errors, modelling pitfalls and that it meets the previously defined functional requirements. Test data coming from a subset of DISNET and automatic association extractions from texts has been transformed according to the proposed ontology to create a Knowledge Graph that can be used in real scenarios, and  which has also been used for the evaluation of the presented ontology.

\end{abstract}
\input{sections/introduction}

\input{sections/relwork}

\input{sections/methodology}

\input{sections/onto-desc}

\input{sections/evaluation}

\input{sections/conclusions}

\subsection*{Acknowledgments}
This work is supported by the DRUGS4COVID++ project, funded by Ayudas Fundación BBVA a equipos de investigación científica SARS-CoV-2 y COVID-19. The work is also supported by “Data-driven drug repositioning applying graph neural networks (3DR-GNN)” under grant “PID2021-122659OB-I00” from the Spanish Ministerio de Ciencia, Innovación y Universidades. LPS’s work is supported by ``Programa de fomento de la investigación y la innovación (Doctorados Industriales)" from Comunidad de Madrid (grant ``IND2019/TIC-17159"). 

%
%
%
\bibliographystyle{splncs04}
\bibliography{references}

\end{document}

%% file: sections/introduction.tex
\section{Introduction}
\label{sec:intro}

The availability of biomedical data has increased in recent decades~\cite{bodenreider_bio-ontologies_2006}. This type of content, whether structured (i.e. relational databases) or unstructured (i.e. text), is usually organized in separate isles owned by companies or institutions, sometimes with proprietary formats. This heterogeneity makes it difficult to extract knowledge through them. The search for drugs, for instance, that could interact with a certain drug, e.g. Plaquenil, during the treatment of COVID-19 based on the experiments published in scientific publications becomes a challenging task. Articles do not usually mention the trade name of drugs, but the active ingredient, Hydroxychloroquine in this case for Plaquenil. The need for massive semantic integration of such information and the establishment of standards, as well as the inclusion of its explicit provenance, is becoming increasingly noticeable. Among the many utilities that they have, bioinformaticians have seen ontologies as a way to manage this explosion of data, facilitating both manual and automatic computer handling. Thus, in the last decades, there has been a considerable adoption of ontologies to model biomedical knowledge~\cite{bodenreider_biomedical_2005}.

Several efforts have been done to integrate this biomedical knowledge in a unique and shared space of representation~\cite{bodenreider_unified_2004,jackson_obo_2021}. One of the most recent works is DISNET\footnote{\url{https://disnet.ctb.upm.es/}}, which provides researches with a platform that enables the creation of complex multilayered graphs~\cite{lagunes_garcia_disnet_2020} following the concepts of Human Disease Networks (HDNs)~\cite{goh_human_2007}. This system aims to give better insights into disease understanding~\cite{prieto_santamaria_classifying_2021,garcia_del_valle_dismanet_2021} and generate new drug repurposing hypotheses~\cite{prieto_santamaria_data-driven_2021,prieto_santamaria_integrating_2022}, by putting together in an accessible knowledge base heterogeneous information that includes large-scale biomedical data obtained and integrated from both structured and unstructured sources. The information in DISNET relational database is organized in three topological levels: (i) the phenotypical (for diseases and their associated symptoms), (ii) the biological (for molecular-shifted data related to
diseases including genes, proteins, metabolic pathways, genetic variants,
non-coding RNAs, etc.) and (iii) the pharmacological (for drugs, their interactions and their connections to diseases).

Despite the efforts of building a unifying and complete biomedical resource, these resources sometimes lack traceability. Users using those resources need to know where each piece of information comes from, and what evidence is supporting it. Thus, completing the resources with this metadata can greatly improve decision making, which is particularly important in the biomedical domain. 

In the current work, we present EBOCA, an ontology that aims to model Evidences for BiOmedical Concepts Association. It describes in two modules (i) biomedical concepts and their associations and (ii) the evidences supporting the associations with metadata and provenance. With this ontology we conceptualize the model that will allow to create a complete semantic resource based on the DISNET biomedical knowledge and associations extracted from texts and other sources enriched with metadata about the evidences supporting them. 

The rest of the paper is organized as follows: Section \ref{sec:rel-work} describes the previous works that have been carried out in the context of the ontology. Section \ref{sec:method} explains the methodology that has been followed to develop the ontology. Section \ref{sec:onto-desc} describes the ontology and its modules in detail, while section \ref{sec:evaluation} develops the evaluation of the ontology. Finally, section \ref{sec:conclusions} draws the conclusions.

%% file: sections/relwork.tex
\section{Related Work}
\label{sec:rel-work}

We divide this section to present similar works related to each of the two modules of EBOCA. That is, we first describe the works that have been previously carried out in the context of biomedical ontologies and secondly include the main ones related to handling evidences, metadata and provenance.

On the one hand, the ontologies that have been proposed to tackle the integration of the biomedical knowledge are multiple, diverse, and different depending on their specific context. Generally, the purpose of biomedical ontologies is to study classes of entities which are of biomedical significance in order to enable sharing complex biological information and the integration of heterogeneous databases \cite{bodenreider_bio-ontologies_2006}. Open Biomedical Ontologies (OBO)~\cite{jackson_obo_2021} was created as an ontology information resource, which now comprises more than 60 ontologies, accessible through BioPortal\footnote{\url{https://bioportal.bioontology.org/}}. Between the different resources that it offers, one can find for instance the Human Phenotype Ontology (HPO)~\cite{kohler_human_2021}, the Sequence Ontology (SO)~\cite{eilbeck_sequence_2005}, or the PRotein Ontology (PRO)~\cite{natale_protein_2011}. OBO Foundry is supported by the National Center for Biomedical Ontology (NCBO) and has a version that is being developed by the National Cancer Institute thesaurus (NCIt)\footnote{\url{https://ncithesaurus.nci.nih.gov/}}. The NCIt provides definitions, synonyms, and other information on nearly 10,000 cancers and related diseases, 17,000 single agents and related substances, and other topics related to cancer and biomedical research.  Also maintained by the USA National Institutes of Health (NIH), in particular, by the National Library of Medicine (NLM), the Unified Medical Language System (UMLS)\footnote{\url{https://uts.nlm.nih.gov/uts/}} is a repository that brings together a great number of health and biomedical vocabularies to enable interoperability~\cite{bodenreider_unified_2004}. In this line, Bio2RDF\footnote{\url{https://bio2rdf.org/}} is a system that uses Semantic Web technologies to build and provide a network of Linked Data for the life sciences~\cite{belleau_bio2rdf_2008}. Regarding the concept of diseases, one of the most renowned efforts to model such an entity resulted in the Disease Ontology (DO)\footnote{\url{https://disease-ontology.org/}}, which has been developed as a standard providing descriptions of human disease terms, phenotype characteristics and related medical vocabulary disease concepts~\cite{schriml_human_2022}. In the specific case of rare diseases, Orphanet\footnote{\url{https://www.orpha.net/}}  developed Orphanet Rare Disease Ontology (ORDO)~\cite{vasant_ordo_2014}. Moreover, other resources have centered their scope in other biomedical entities. DisGeNET integrates associations between genes and variants to human diseases~\cite{pinero_disgenet_2020}. In the context of associations between biomedical entities, the Semanticscience Integrated Ontology (SIO) provides a simple, integrated ontology of types and relations for rich description of objects, processes and their attributes~\cite{dumontier_semanticscience_2014}. As for proteins, other databases and terminologies include UniProt~\cite{redaschi_uniprot_2009} or NeXtProt~\cite{zahn-zabal_nextprot_2020}. WikiPathways represents and integrated data regarding biological pathways~\cite{martens_wikipathways_2021}. Finally, drugs and their related information (e.g. their interactions) have been modeled and integrated in several sources. Between those, we can find ChEMBL~\cite{mendez_chembl_2019}, the Comparative Toxicogenomics Database (CTD)~\cite{davis_comparative_2021}, DrugBank~\cite{wishart_drugbank_2018} or TWOSIDES\footnote{\url{https://nsides.io/}}.

On the other hand, not so many ontologies include evidence information. With the exponential adoption of computational techniques able to discover new knowledge, the need to track the evidence and provenance of these techniques is becoming increasingly important. The Evidence and Conclusion Ontology (ECO)~\cite{giglio2019eco} was developed to tackle this issue. It provides a controlled vocabulary that enables describing the type of evidence of an assertion, with a focus on the biomedical domain. This ontology is maintained and curated with active participation from the community, as it is currently used by projects such as the Gene Ontology (GO)~\cite{gene2004gene} and DisGeNET~\cite{pinero_disgenet_2020}. To track evidences, more metadata is required, and many ontologies have been developed for this purpose and are widely used. Some examples are the PROV Ontology (PROV-O)~\cite{lebo2013prov}, its extension Provenance, Authoring and Versioning (PAV)~\cite{ciccarese2013pav}, and DCMI Metadata Terms\footnote{\url{https://www.dublincore.org/specifications/dublin-core/dcmi-terms/}}. The module EBOCA Evidences reuses entities from these well-stablished ontologies to provide provenance and metadata of biomedical associations evidences, that may come from both curated data sources or inferred from unstructured texts.

%% file: sections/methodology.tex
\section{Methodology}
\label{sec:method}

\begin{figure}[!t]
\centering
\includegraphics[width=0.65\linewidth]{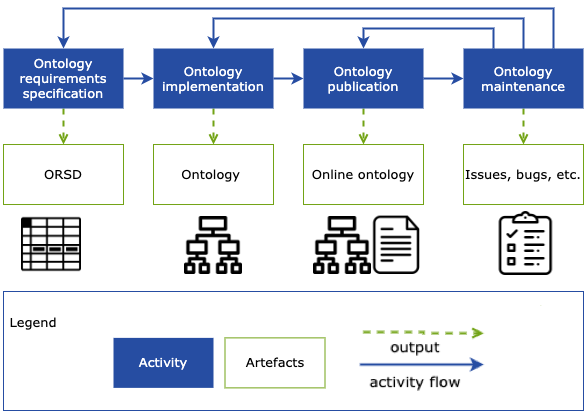}
\caption{LOT Methodology~\cite{POVEDAVILLALON2022104755}}
\label{fig:lot}
\end{figure}

The EBOCA Ontology was developed following the guidelines provided by the Linked Open Terms (LOT) methodology~\cite{POVEDAVILLALON2022104755}. 
LOT, based on the NeOn methodology~\cite{suarez2015neon}, is a lightweight methodology for ontology and vocabulary development. 
It includes four major stages: Requirements Specification, Implementation, Publication, and Maintenance (Figure \ref{fig:lot}). 
In this section, we describe these stages and how they have been applied and adapted to the development of EBOCA.

\textbf{Requirements specification.} This first stage involves the activities that lead to defining the requirements that the ontology must meet. Those are the purpose and scope of the ontology, i.e., the objective of the ontology and its extent, the knowledge it models; and the requirements specified in the form of competency questions and/or affirmative statements. Both purpose and scope are specified in the ontology documentation. The requirements can be found in the repositories of each module, and accessed through the ontology portal~\footnote{\url{https://w3id.org/eboca/portal}}.

\textbf{Implementation.} The goal of this stage is to build the ontology using a formal language, using the requirements identified in the previous stage as guidance. The first version of the ontology is conceptualized based on these requirements, and subsequently refined by verifying the model with domain experts. The conceptualization is carried out representing the ontology graphically following the Chowlk notation~\cite{chavez2022chowlk} and implemented in OWL 2 with Protégé. The evaluation of the ontology is carried out using (i) SPARQL queries and Themis~\cite{fernandez2021themis} to validate the requirements, (ii) OOPS!~\cite{poveda2014oops} to identify modelling pitfalls and (iii) the Pellet and HermiT reasoners to check for inconsistencies.

\textbf{Publication.} The publication stage refers to the activities carried out to make the ontology and its documentation available. The documentation is generated with Widoco~\cite{garijo2017widoco}, and it is published with a W3ID URI\footnote{\url{https://w3id.org/eboca/sem-disnet},\url{https://w3id.org/eboca/evidences}}. The published EBOCA resources can be accessed through the ontology portal. 

\textbf{Maintenance.} Finally, the maintenance stage aims to ensure that the ontology is updated with error corrections and new requirements. The EBOCA ontology enables the gathering of new requirements and issues through GitHub and GitLab repositories. 

%% file: sections/onto-desc.tex
\section{Ontology description}
\label{sec:onto-desc}

The EBOCA Ontology describes the evidences of associations between biomedical concepts. It is composed of two parts, one for modelling biomedical concepts and associations, EBOCA SEM-DISNET; and the other for representing the evidences of these associations with metadata and provenance information, EBOCA Evidences. They are designed and developed as separated modules with one class of connection, \texttt{sio:SIO\_000897} (``Association''), that will be described further in this section. All related resources of this ontology are publicly available online~\footnote{\url{https://w3id.org/eboca/portal}}.

\subsection{EBOCA SEM-DISNET}

The EBOCA SEM-DISNET module\footnote{\url{https://w3id.org/eboca/sem-disnet}} is designed to represent the associations of common biomedical concepts. These concepts include principally: diseases, phenotypes, genes, genetical variants, biological pathways, drugs, proteins and targets. The associations link pairs of concepts, e.g. gene-disease or drug-disease association. This module is also focused on representing the data contained in DISNET relational database ~\cite{lagunes_garcia_disnet_2020} and give them a semantic structure. These data are organized in three layers: (i) the phenotypical (in which disease-phenotype relationships are gathered from applying text mining processes to Wikipedia, PubMed and MayoClinic), the biological (with diseases and their relationships to genes, proteins, genetical variants, pathways and so on, extracted from structured sources as DisGeNET) and the pharmacological (which stores drugs and drug-related information from structured sources as ChEMBL).

The requirements of this module are formalized as 15 competency questions, validated through SPARQL queries (see Section \ref{sec:evaluation}). These functional requirements make reference to the scope of the ontology, which has been defined as the already mentioned biomedical concepts as well as the association between those concepts. The final intended users of EBOCA SEM-DISNET could mainly be research scientists in the field of life science and computer science, and some of the most direct uses would be the aforementioned generation of drug repurposing hypotheses or the obtainment of a better understanding of diseases.

This module of EBOCA is mostly built from the reuse of a wide amount of existing ontologies. That is, the majority of its entities have been obtained from previously developed terminologies, when possible. The creation of new concepts from scratch has only been implemented when it was totally unavoidable. The general design pattern used to model this module is based on DisGeNET~\cite{queralt_rosinach_disgenet_rdf_2016}, that provides RDF resources integrating information of associations between genes and diseases (and others such as variants and diseases). This design pattern was suitable enough to represent other types of associations as well. We considered that the details and representations established by this patterns were in accordance with the information to be modeled from the other named associaditons.  EBOCA SEM-DISNET consists of a total of 29 classes and 33 object properties reused from different previously published vocabularies. However, in some occasions, it was not accurate to reuse certain classes due to their absence or because, even if they existed, they did not fit the specifications of concept to be modeled. This was the case of 5 classes and 12 object properties, which have been specifically created for EBOCA SEM-DISNET. 

On the one hand, for both the classes modeling concepts and associations, most terms have been reused from NCIt, closely followed by SIO, and to a lesser extent by ChEMBL and ORDO. Furthermore, the classes corresponding to associations between concepts  were incorporated into a class hierarchy, and was represented independently of the main diagram for the sake of clarity. In this hierarchy, all association entities were modeled as subclasses of the \texttt{sio:SIO\_000897} class, denoting ``Association'' and chosen to link both EBOCA modules. It can also occur that the reused class is not directly related to the ``Association'' class. In these cases, for example, for the \texttt{cco:Mechanism} class, the strategy consisted in modelling a SEM-DISNET class named \texttt{DrugTargetAssociation}, which is a subclass of ChEMBL's \texttt{cco:Mechanism} and ``Association''.

On the other hand, with respect to properties, it was intended that the relationships between classes were as homogeneous as possible. However, while most have been reused from SIO (almost half of them), it has been necessary to reuse some from ChEMBL and the NCIt. SEM-DISNET relationships were created with the purpose of joining the meaning of some of the reused object properties with \texttt{sio:SIO\_000628} (which means `refers to') or \texttt{sio:SIO\_000212} (`is referred to by'). We followed the same strategy as for the classes. An object property was modeled under the SEM-DISNET namespace and classified as a subproperty of the mentioned `refers to' or `is referred to by' according to its meaning, and the one from the original ontology, establishing a hierarchy\footnote{\url{https://medal.ctb.upm.es/internal/gitlab/disnet/sem-disnet/blob/master/diagrams/SEM-DISNET\_hierarchy.png}}. For example, we created the object property \texttt{drugForMechanism} which is at the same time subproperty of \texttt{cco:hasMechanism} and \texttt{sio:SIO\_000212}. 

EBOCA SEM-DISNET revolves around the concept of disease, modeled in \texttt{ncit:C7057} class, reused from the NCIt. The semantic type is specified by the class \texttt{sio:SIO\_000326}, while the disease class (meaning categorizations of diseases) is defined under \texttt{obo:HP\_0000118}. Disease markers are modeled with \texttt{ncit:C18329}. The \texttt{ncit:C7057} class is associated to classes that represent relationships between concepts, including the associations with non-coding RNAs (\texttt{ncit:C26549}) and with Orphanet classification \texttt{ordo:Orphanet\_557492} associations. Other classes modeling the relationships with diseases include:

\begin{figure}[!t]
\centering
\includegraphics[width=1\linewidth]{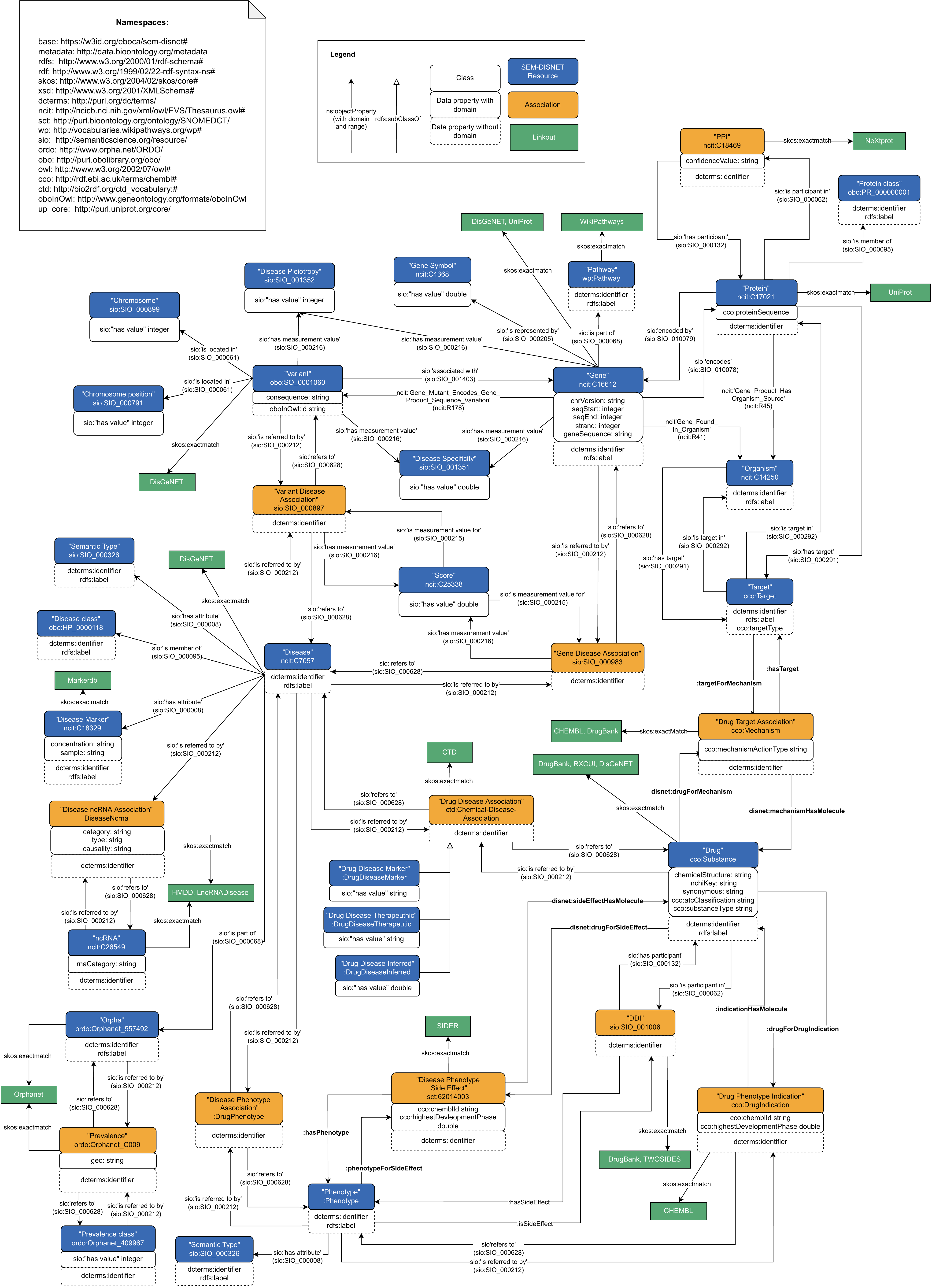}
\caption{Conceptualization of the EBOCA SEM-DISNET module.}
\label{fig:ebocadisnet}
\end{figure}

\begin{itemize}
    \item \textbf{Disease-gene association.} Modeled with \texttt{sio:SIO\_000983}, it is quantified by the NCIt class \texttt{ncit:C25338} (labeled `Score'). The \texttt{ncit:C16612} class (labeled `Gene'), is related with WikiPathways metabolic \texttt{wp:Pathway}, the associated \texttt{obo:SO\_0001060} (`Variant') and the encoded \texttt{ncit:C17021} (`Protein'). Each of these proteins belongs to an \texttt{obo:PR\_000000001} (referring to a `Protein class') and participates in \texttt{ncit:C18469} (
   `Protein-Protein Interactions' or PPIs). Both proteins and genes are related to the class \texttt{ncit:C14250} (`Organism') and proteins and organisms can act as a \texttt{cco:Target}, class reused from the ChEMBL ontology.
    \item \textbf{Disease-variant association.} It has an associated score defined by the mentioned `Score' class. The \texttt{obo:SO\_0001060} class (representing `Variant') is reused from OBO Sequence Ontology (SO).
    \item \textbf{Disease-phenotype association.} This association and the related class \texttt{Pheno\-type} have been modeled by SEM-DISNET, since no classes were found in other ontologies with the exact meaning and specifications corresponding to the one in DISNET database.
    \item \textbf{Disease-drug association.} The association between diseases and drugs was modeled reusing a class from CTD, \texttt{ctd:Chemical-Disease-Associati\-on}. Drug-disease pairs were classified into three created subclasses depending on the effect that the drug has on the disease. It can act as a marker if the chemical correlates with a disease (\texttt{DrugDiseaseMarker}), it can be used to treat the disease (\texttt{DrugDiseaseTherapeutic}), or the association might have been inferred (\texttt{DrugDiseaseInferred}). 

\end{itemize}

The \texttt{cco:Drug} class was reused from the ChEMBL ontology and represents any substance which when absorbed into a living organism may modify one or more of its functions. Besides its associated properties, it references to classes which represent relationships:
\begin{itemize}
    \item \textbf{Drug-target association.} Corresponding to \texttt{cco:Mechanism}, it is known as the mechanism of action of a drug when it addresses to \texttt{cco:Target}. 
    \item \textbf{Drug-phenotype association.} The two entities \texttt{cco:Drug} and \texttt{Phenotype} are connected by three association classes referring to the possible causality of the interaction:
    \begin{itemize}
        \item \textbf{Indication.} Reused from ChEMBL class \texttt{cco:DrugIndication}, it is referred to the approved uses of a medication.
        \item \textbf{Side effect.} Association between symptoms considered as adverse effects that cause patient phenotype changes in response to the treatment with a drug. It was taken from SNOMED CT database (\texttt{sct:662014003}).
        \item \textbf{Drug-drug interaction (DDI).} The class \texttt{sio:SIO\_001006} was reused to model these interactions. DrugBank is the main source of data providing DDIs in DISNET, but their redistribution is not allowed. However, DISNET also holds TWOSIDES information, which also provides the potential adverse side effects that the DDI might cause.
    \end{itemize}
\end{itemize}
EBOCA SEM-DISNET ontological model is represented in Figure \ref{fig:ebocadisnet}. The resource classes, represented by blue rounded boxes, correspond to biomedical entities; while association classes, which establish the relationship between two biomedical entities, are represented in orange. Attributes correspond to the white rounded boxes, and linkouts to external original sources to green boxes.

\subsection{EBOCA Evidences}

The EBOCA Evidences module\footnote{\url{https://w3id.org/eboca/evidences}} is built to enrich the EBOCA SEM-DISNET module. It is focused on providing metadata and provenance information about the associations between biomedical concepts. These evidences of associations may come from well known curated sources, or may be extracted or inferred from texts. This module enables describing in more detail the type of evidence supporting the association, the agents involved in its extraction and publication, and if applicable, the texts and sources where the evidence is extracted from.

This module reuses the following ontologies: SIO~\cite{dumontier_semanticscience_2014} for associations; Evidence \& Conclusion Ontology (ECO)~\cite{giglio2019eco} for the different kinds of evidences; Friend Of A Friend (FOAF)~\cite{graves2007foaf} for creators of evidences; DCMI Metadata Terms\footnote{\url{https://www.dublincore.org/specifications/dublin-core/dcmi-terms/}} for general metadata; Provenance, Authoring and Versioning (PAV)~\cite{ciccarese2013pav} for metadata of evidences; and the SPAR Ontologies~\cite{peroni2018spar} FRBR-aligned Bibliographic Ontology (FaBiO), Document Components Ontology (DoCO) to describe and expression of work and its parts, paragraphs.


\begin{figure}[!t]
\centering
\includegraphics[width=0.85\linewidth]{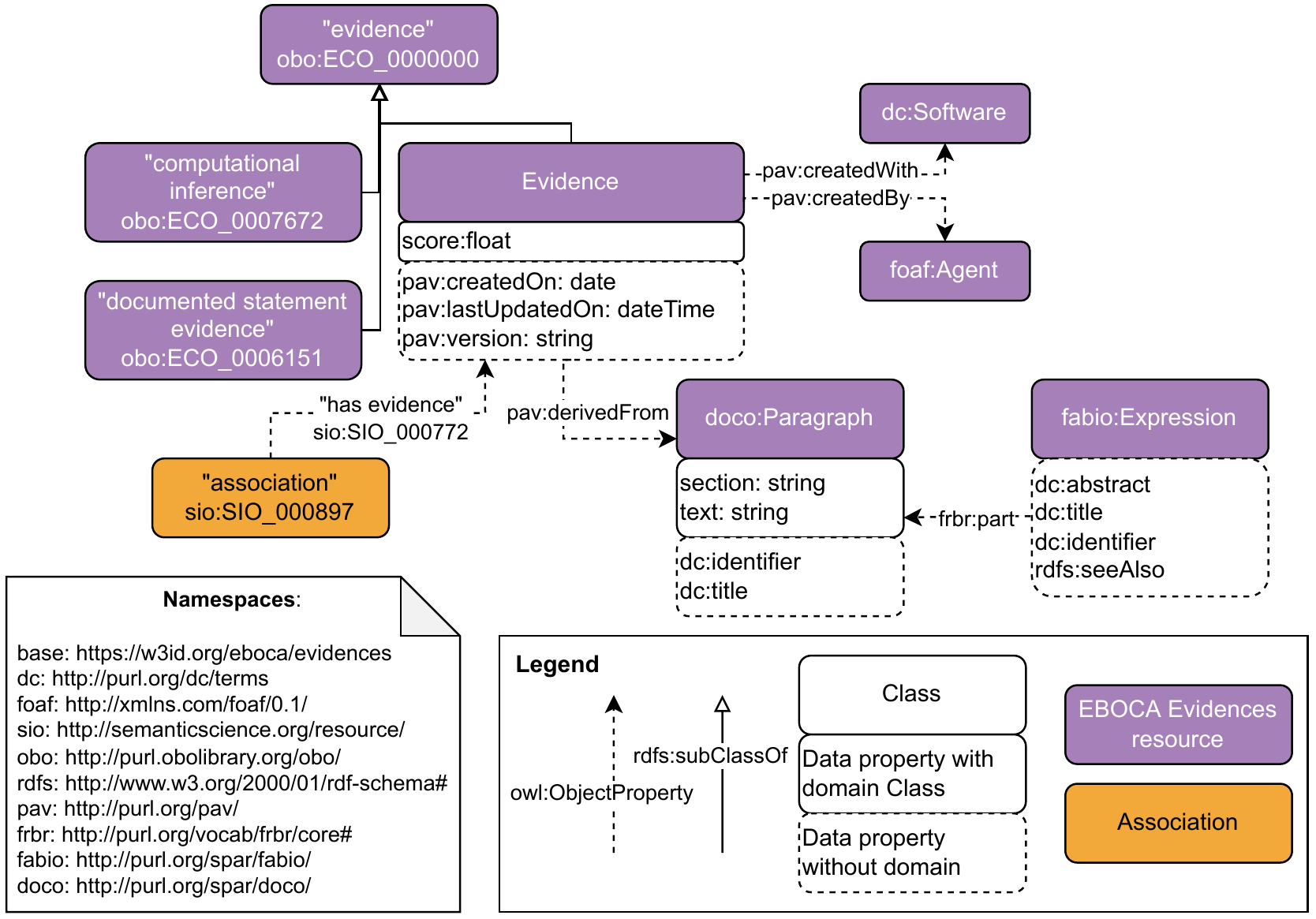}
\caption{Conceptualization of the EBOCA Evidences module.}
\label{fig:ebocaev}
\end{figure}

The conceptualization of the EBOCA Evidences module is depicted in Figure \ref{fig:ebocaev}. The link to the SEM-DISNET module is the class  \texttt{sio:SIO\_000897} (``Association''), which is the superclass of all kinds of associations represented in the other module. These associations may have a supporting \texttt{Evidence}, which in turn can be inferred (\texttt{ECO:0007672} ``computational inference'') or extracted from curated sources (\texttt{ECO:0006151} ``documented statement evidence''). The class \texttt{Evidence} also enables describing the date of creation and update, version, software and agent responsible for its creation, and its provenance. 

This module is focused on describing paragraphs from curated resources and natural text, as it is the main use case. Thus, this module represents how evidences are derived from (\texttt{pav:derivedFrom}) \texttt{doco:Paragraph}, which is part of \texttt{fabio:Expression}, a class that describes, among others, papers. Both paragraphs and expressions can be uniquely identified, and include more attributes to describe them (e.g. the section paragraphs belong to and complete text; title, abstract and URL for papers). Evidences can also be derived from other sources, case in which the resource URI is specified with \texttt{pav:derivedFrom} too.

The scope of this module is limited for now to describe provenance of evidences when extracted from texts and curated resources. As the EBOCA resource grows and adds more different kinds of evidence extractions methods, it will evolve to represent them. The specific requirements for EBOCA Evidences are publicly available\footnote{\url{https://drugs4covid.github.io/EBOCA-portal/requirements/requirements-evidences.html}}. These requirements are tested in Themis and with SPARQL queries, as Section \ref{sec:evaluation} explains in detail.

%% file: sections/evaluation.tex
\section{Ontology Evaluation}
\label{sec:evaluation}

Both EBOCA modules are evaluated in different ways to identify inconsistencies, pitfalls, errors and to check that they meet the requirements.

The modelling pitfalls were evaluated with the OntOlogy Pitfall Scanner! (OOPS!)~\cite{poveda2014oops}. As many ontologies are reused in both modules, some pitfalls appear, but all of them pointing to reused ontology entities. The pitfall record includes P04 (Creating unconnected ontology elements), P08 (Missing annotations), P11 (Missing domain or range in properties), P13 (Inverse relationships not explicitly declared) and P22 (Using different naming conventions in the ontology). However, all pitfalls that appeared in the modules for newly implemented entities during the implementation have been solved. Inconsistencies were checked with the reasoners Pellet and HermiT. For both modules, no errors were found when running the mentioned reasoners. To ensure that the modules met the requirements, test data was transformed into RDF, building a small Knowledge Graph (KG), which was used to run the corresponding SPARQL queries. This transformation was performed separately in each module. 

To build each KG, declarative mapping languages are used: the W3C Recommendation R2RML~\cite{das2012r2rml} for EBOCA SEM-DISNET, and RML~\cite{Dimou2014rml} for EBOCA Evidences. Both mappings have been created using Mapeathor~\cite{iglesias2020mapeathor}. The mapping engine Morph-KGC~\cite{arenas2022morph-kgc} has been used to materialize both KGs. The resulting KGs and related resources (mappings, queries) are available in GitHub\footnote{\url{https://github.com/drugs4covid/EBOCA-Resources}}. The mappings can also be used for virtualization, but for evaluating the ontology we chose the materialization approach (i.e. generating the RDF triples).

For EBOCA SEM-DISNET, as the original relational database scales to large amounts of data, we decided to use a subset of the entirety of DISNET information to generate the KG. We included several concepts and some of their associations. Those concepts were diseases, genes, drugs and pathways. As for the associations, the data corresponding to disease-gene, disease-variant, disease-drug, gene-variant, gene-pathway and drug-drug associations were transformed to RDF. Despite the fact that it was just a fragment, this part of the KG contained 8,691,974 triplets, weighing a total of 1.6 GB.
The graph was uploaded to Blazegraph\footnote{\url{https://blazegraph.com/}}, which enabled the successful execution of 15 SPARQL queries. They represent the different competency questions, 
validating the variety of concepts and associations in EBOCA SEM-DISNET module. The queries were targeted to both the modeled ontology as well as to the RDF instances (named individuals) created and included in the KG.

For EBOCA Evidences, JSON data coming from biomedical concepts extracted from 5,000 paragraphs of the CORD-19 corpus were used to build the test KG. A Named Entity Recognition and Normalization system (BioNER+BioNEN), based on fine-tuned BioBERT models~\cite{badenes2022bioner}, was used to identify diseases, drugs and genetic terms. Thus, associations from the extracted terms were transformed and annotated with metadata about their evidence and provenance. The queries were run in Blazegraph with success. The exception was query \texttt{eboca-ev7}, that cannot output any results because of the extraction method used outputs no confidence score. However, to check the ontology consistency independently of the test data, the requirements formulated as statements were also validated using Themis~\cite{fernandez2021themis}. Themis is a web service that enables executing tests to ontologies. These tests correspond to the ontology requirements, have been tested with success. 

Evaluations for both modules were run successfully, with good results from OOPS!, SPARQL queries and Themis. This ensures that the ontology is consistent, meets the requirements, and to the best of our knowledge, has no errors.

%% file: sections/conclusions.tex
\section{Conclusions}
\label{sec:conclusions}

This work presents EBOCA, an ontology that models evidences and provenance of associations between biomedical concepts. It is composed of two modules, (i) EBOCA SEM-DISNET for biomedical concepts and associations, and (ii) EBOCA Evidences for evidence, provenance and metadata information of associations. EBOCA aims to put forward a resource to model biomedical concepts and their association with the possibility of tracing them via evidences. We have explained LOT, the methodology employed for the development of these two modules; we have described them, and how their evaluation was performed to validate them. We conclude that the proposed ontology is fit to its purpose, to represent biomedical entities, their associations and evidence supporting them.

As future work, we want to create a complete Knowledge Graph to integrate data from DISNET enriched with its evidence information, as well as new associations from unstructured text analysis. Moreover, the evidences will be published as Nanopublications to promote its reuse. Other applications we foresee from the use of EBOCA are the possibility of proposing drug candidates for new diseases by approaching the link prediction challenge in the graph, addressing a better understanding of diseases by means of knowledge reasoning, or detecting and managing contradictory evidences.

%% file: main.bbl
\begin{thebibliography}{10}
\providecommand{\url}[1]{\texttt{#1}}
\providecommand{\urlprefix}{URL }
\providecommand{\doi}[1]{https://doi.org/#1}

\bibitem{arenas2022morph-kgc}
Arenas-Guerrero, J., Chaves-Fraga, D., Toledo, J., Pérez, M.S., Corcho, O.:
  {Morph-KGC: Scalable Knowledge Graph Materialization with Mapping
  Partitions}. Semantic Web  (2022),
  \url{http://www.semantic-web-journal.net/system/files/swj3135.pdf}

\bibitem{badenes2022bioner}
Badenes-Olmedo, C., Alonso, A., Corcho, O.: An overview of drugs, diseases,
  genes and proteins in the cord-19 corpus. Procesamiento del Lenguaje Natural
  \textbf{69} (2022)

\bibitem{belleau_bio2rdf_2008}
Belleau, F., Nolin, M.A., Tourigny, N., Rigault, P., Morissette, J.: Bio2rdf:
  Towards a mashup to build bioinformatics knowledge systems. Journal of
  Biomedical Informatics  \textbf{41}(5),  706--716 (2008).
  \doi{10.1016/j.jbi.2008.03.004}

\bibitem{bodenreider_unified_2004}
Bodenreider, O.: The {Unified} {Medical} {Language} {System} ({UMLS}):
  integrating biomedical terminology. Nucleic Acids Research
  \textbf{32}(suppl\_1),  D267--D270 (Jan 2004). \doi{10.1093/nar/gkh061}

\bibitem{bodenreider_biomedical_2005}
Bodenreider, O., Mitchell, J.A., McCray, A.T.: Biomedical ontologies. In:
  Pacific Symposium on Biocomputing. pp. 76--78 (2005)

\bibitem{bodenreider_bio-ontologies_2006}
Bodenreider, O., Stevens, R.: Bio-ontologies: current trends and future
  directions. Briefings in Bioinformatics  \textbf{7}(3),  256--274 (2016).
  \doi{10.1093/bib/bbl027}

\bibitem{chavez2022chowlk}
Chávez-Feria, S., García-Castro, R., Poveda-Villalón, M.: Chowlk: from
  uml-based ontology conceptualizations to owl. In: Groth, P., Vidal, M.E.,
  Suchanek, F., Szekely, P., Kapanipathi, P., Pesquita, C., Skaf-Molli, H.,
  Tamper, M. (eds.) The Semantic Web. pp. 338--352. Springer International
  Publishing, Cham (2022)

\bibitem{ciccarese2013pav}
Ciccarese, P., Soiland-Reyes, S., Belhajjame, K., Gray, A.J., Goble, C., Clark,
  T.: Pav ontology: provenance, authoring and versioning. Journal of biomedical
  semantics  \textbf{4}(1),  1--22 (2013)

\bibitem{gene2004gene}
Consortium, G.O.: The gene ontology ({GO}) database and informatics resource.
  Nucleic Acids Research  \textbf{32}(suppl\_1),  D258--D261 (2004)

\bibitem{das2012r2rml}
Das, S., Sundara, S., Cyganiak, R.: {R2RML: RDB to RDF Mapping Language, W3C
  Recommendation 27 September 2012}. www.w3.org/TR/r2rml  (2012)

\bibitem{davis_comparative_2021}
Davis, A.P., Grondin, C.J., Johnson, R.J., Sciaky, D., Wiegers, J., Wiegers,
  T.C., Mattingly, C.J.: Comparative toxicogenomics database ({CTD}): update
  2021. Nucleic Acids Research  \textbf{49},  D1138--D1143 (2021).
  \doi{10.1093/nar/gkaa891}

\bibitem{Dimou2014rml}
Dimou, A., Sande, M.V., Colpaert, P., Verborgh, R., Mannens, E., {Van De
  Walle}, R.: {RML: A generic language for integrated RDF mappings of
  heterogeneous data}. In: LDOW (2014)

\bibitem{dumontier_semanticscience_2014}
Dumontier, M., Baker, C.J., Baran, J., Callahan, A., Chepelev, L., Cruz-Toledo,
  J., Del~Rio, N.R., Duck, G., Furlong, L.I., Keath, N., Klassen, D.,
  {McCusker}, J.P., Queralt-Rosinach, N., Samwald, M., Villanueva-Rosales, N.,
  Wilkinson, M.D., Hoehndorf, R.: The semanticscience integrated ontology
  ({SIO}) for biomedical research and knowledge discovery. Journal of
  Biomedical Semantics  \textbf{5}(1), ~14 (2014). \doi{10.1186/2041-1480-5-14}

\bibitem{eilbeck_sequence_2005}
Eilbeck, K., Lewis, S.E., Mungall, C.J., Yandell, M., Stein, L., Durbin, R.,
  Ashburner, M.: The {Sequence} {Ontology}: a tool for the unification of
  genome annotations. Genome Biology  \textbf{6}(5), ~R44 (Apr 2005).
  \doi{10.1186/gb-2005-6-5-r44}

\bibitem{fernandez2021themis}
Fern{\'a}ndez-Izquierdo, A., Cimmino, A., Garc{\'\i}a-Castro, R.: Supporting
  demand-response strategies with the delta ontology. In: 2021 IEEE/ACS 18th
  International Conference on Computer Systems and Applications (AICCSA).
  pp.~1--8. IEEE (2021)

\bibitem{garijo2017widoco}
Garijo, D.: Widoco: a wizard for documenting ontologies. In: International
  Semantic Web Conference. pp. 94--102. Springer (2017)

\bibitem{giglio2019eco}
Giglio, M., Tauber, R., Nadendla, S., Munro, J., Olley, D., Ball, S., Mitraka,
  E., Schriml, L.M., Gaudet, P., Hobbs, E.T., et~al.: Eco, the evidence \&
  conclusion ontology: community standard for evidence information. Nucleic
  Acids Research  \textbf{47}(D1),  D1186--D1194 (2019)

\bibitem{goh_human_2007}
Goh, K.I., Cusick, M.E., Valle, D., Childs, B., Vidal, M., Barabási, A.L.: The
  human disease network. Proceedings of the National Academy of Sciences
  \textbf{104}(21),  8685--8690 (May 2007). \doi{10.1073/pnas.0701361104}

\bibitem{graves2007foaf}
Graves, M., Constabaris, A., Brickley, D.: Foaf: Connecting people on the
  semantic web. Cataloging \& classification quarterly  \textbf{43}(3-4),
  191--202 (2007)

\bibitem{iglesias2020mapeathor}
Iglesias-Molina, A., Pozo-Gilo, L., Do{\c{n}}a, D., Ruckhaus, E., Chaves-Fraga,
  D., Corcho, {\'O}.: Mapeathor: Simplifying the specification of declarative
  rules for knowledge graph construction. In: ISWC (Demos/Industry) (2020)

\bibitem{jackson_obo_2021}
Jackson, R., Matentzoglu, N., Overton, J.A., Vita, R., Balhoff, J.P.,
  Buttigieg, P.L., Carbon, S., Courtot, M., Diehl, A.D., Dooley, D.M., Duncan,
  W.D., Harris, N.L., Haendel, M.A., Lewis, S.E., Natale, D.A.,
  Osumi-Sutherland, D., Ruttenberg, A., Schriml, L.M., Smith, B.,
  Stoeckert~Jr., C.J., Vasilevsky, N.A., Walls, R.L., Zheng, J., Mungall, C.J.,
  Peters, B.: {OBO} foundry in 2021: operationalizing open data principles to
  evaluate ontologies. Database  \textbf{2021},  baab069 (2021).
  \doi{10.1093/database/baab069}

\bibitem{kohler_human_2021}
Köhler, S., Gargano, M., Matentzoglu, N., Carmody, L.C., Lewis-Smith, D.,
  Vasilevsky, N.A., Danis, D., Balagura, G., Baynam, G., Brower, A.M.,
  Callahan, T.J., Chute, C.G., Est, J.L., Galer, P.D., Ganesan, S., Griese, M.,
  Haimel, M., Pazmandi, J., Hanauer, M., Harris, N.L., Hartnett, M.,
  Hastreiter, M., Hauck, F., He, Y., Jeske, T., Kearney, H., Kindle, G., Klein,
  C., Knoflach, K., Krause, R., Lagorce, D., McMurry, J.A., Miller, J.A.,
  Munoz-Torres, M., Peters, R.L., Rapp, C.K., Rath, A.M., Rind, S.A.,
  Rosenberg, A., Segal, M.M., Seidel, M.G., Smedley, D., Talmy, T., Thomas, Y.,
  Wiafe, S.A., Xian, J., Yüksel, Z., Helbig, I., Mungall, C.J., Haendel, M.A.,
  Robinson, P.N.: The {Human} {Phenotype} {Ontology} in 2021. Nucleic Acids
  Research  \textbf{49}(D1),  D1207--D1217 (Jan 2021).
  \doi{10.1093/nar/gkaa1043}

\bibitem{lagunes_garcia_disnet_2020}
Lagunes-García, G., Rodríguez-González, A., Prieto-Santamaría, L., Valle,
  E.P.G.d., Zanin, M., Menasalvas-Ruiz, E.: {DISNET}: a framework for
  extracting phenotypic disease information from public sources. PeerJ
  \textbf{8},  e8580 (Feb 2020). \doi{10.7717/peerj.8580}

\bibitem{lebo2013prov}
Lebo, T., Sahoo, S., McGuinness, D., Belhajjame, K., Cheney, J., Corsar, D.,
  Garijo, D., Soiland-Reyes, S., Zednik, S., Zhao, J.: Prov-o: The prov
  ontology. www.w3.org/TR/prov-o/  (2013)

\bibitem{martens_wikipathways_2021}
Martens, M., Ammar, A., Riutta, A., Waagmeester, A., Slenter, D., Hanspers, K.,
  A. Miller, R., Digles, D., Lopes, E., Ehrhart, F., Dupuis, L.J., Winckers,
  L.A., Coort, S., Willighagen, E.L., Evelo, C.T., Pico, A.R., Kutmon, M.:
  {WikiPathways}: connecting communities. Nucleic Acids Research
  \textbf{49}(D1),  D613--D621 (Jan 2021). \doi{10.1093/nar/gkaa1024}

\bibitem{mendez_chembl_2019}
Mendez, D., Gaulton, A., Bento, A.P., Chambers, J., De Veij, M., Félix, E.,
  Magariños, M., Mosquera, J., Mutowo, P., Nowotka, M., Gordillo-Marañón,
  M., Hunter, F., Junco, L., Mugumbate, G., Rodriguez-Lopez, M., Atkinson, F.,
  Bosc, N., Radoux, C., Segura-Cabrera, A., Hersey, A., Leach, A.: {ChEMBL}:
  towards direct deposition of bioassay data. Nucleic Acids Research
  \textbf{47},  D930--D940 (2019). \doi{10.1093/nar/gky1075}

\bibitem{natale_protein_2011}
Natale, D.A., Arighi, C.N., Barker, W.C., Blake, J.A., Bult, C.J., Caudy, M.,
  Drabkin, H.J., D’Eustachio, P., Evsikov, A.V., Huang, H., Nchoutmboube, J.,
  Roberts, N.V., Smith, B., Zhang, J., Wu, C.H.: The {Protein} {Ontology}: a
  structured representation of protein forms and complexes. Nucleic Acids
  Research  \textbf{39}(suppl\_1),  D539--D545 (Jan 2011).
  \doi{10.1093/nar/gkq907}

\bibitem{peroni2018spar}
Peroni, S., Shotton, D.: The spar ontologies. In: International Semantic Web
  Conference. pp. 119--136. Springer (2018)

\bibitem{pinero_disgenet_2020}
Piñero, J., Ramírez-Anguita, J.M., Saüch-Pitarch, J., Ronzano, F., Centeno,
  E., Sanz, F., Furlong, L.I.: The {DisGeNET} knowledge platform for disease
  genomics: 2019 update. Nucleic Acids Research  \textbf{48},  D845--D855
  (2020). \doi{10.1093/nar/gkz1021}

\bibitem{poveda2014oops}
Poveda-Villal{\'o}n, M., G{\'o}mez-P{\'e}rez, A., Su{\'a}rez-Figueroa, M.C.:
  Oops!(ontology pitfall scanner!): An on-line tool for ontology evaluation.
  International Journal on Semantic Web and Information Systems (IJSWIS)
  \textbf{10}(2),  7--34 (2014)

\bibitem{POVEDAVILLALON2022104755}
Poveda-Villalón, M., Fernández-Izquierdo, A., Fernández-López, M.,
  García-Castro, R.: {LOT: An industrial oriented ontology engineering
  framework}. Engineering Applications of Artificial Intelligence
  \textbf{111},  104755 (2022).
  \doi{https://doi.org/10.1016/j.engappai.2022.104755}

\bibitem{prieto_santamaria_integrating_2022}
Prieto~Santamaría, L., Díaz~Uzquiano, M., Ugarte~Carro, E., Ortiz-Roldán,
  N., Pérez~Gallardo, Y., Rodríguez-González, A.: Integrating heterogeneous
  data to facilitate {COVID}-19 drug repurposing. Drug Discovery Today
  \textbf{27}(2),  558--566 (Feb 2022). \doi{10.1016/j.drudis.2021.10.002}

\bibitem{prieto_santamaria_data-driven_2021}
Prieto~Santamaría, L., Ugarte~Carro, E., Díaz~Uzquiano, M., Menasalvas~Ruiz,
  E., Pérez~Gallardo, Y., Rodríguez-González, A.: A data-driven methodology
  towards evaluating the potential of drug repurposing hypotheses.
  Computational and Structural Biotechnology Journal  \textbf{19},  4559--4573
  (Jan 2021). \doi{10.1016/j.csbj.2021.08.003}

\bibitem{prieto_santamaria_classifying_2021}
Prieto~Santamaría, L., García~del Valle, E.P., Zanin, M., Hernández~Chan,
  G.S., Pérez~Gallardo, Y., Rodríguez-González, A.: Classifying diseases by
  using biological features to identify potential nosological models.
  Scientific Reports  \textbf{11}(1),  21096 (Oct 2021).
  \doi{10.1038/s41598-021-00554-6}

\bibitem{queralt_rosinach_disgenet_rdf_2016}
Queralt-Rosinach, N., Piñero, J., Bravo, A., Sanz, F., Furlong, L.I.:
  {DisGeNET}-{RDF}: harnessing the innovative power of the semantic web to
  explore the genetic basis of diseases. Bioinformatics  \textbf{32}(14),
  2236--2238 (2016)

\bibitem{redaschi_uniprot_2009}
Redaschi, N., Consortium, U.: {UniProt} in {RDF}: Tackling data integration and
  distributed annotation with the semantic web. Nature Precedings pp.~1--1
  (2019). \doi{10.1038/npre.2009.3193.1}

\bibitem{schriml_human_2022}
Schriml, L.M., Munro, J.B., Schor, M., Olley, D., {McCracken}, C., Felix, V.,
  Baron, J., Jackson, R., Bello, S., Bearer, C., Lichenstein, R., Bisordi, K.,
  Dialo, N.C., Giglio, M., Greene, C.: The human disease ontology 2022 update.
  Nucleic Acids Research  \textbf{50},  D1255--D1261 (2022).
  \doi{10.1093/nar/gkab1063}

\bibitem{suarez2015neon}
Su{\'a}rez-Figueroa, M.C., G{\'o}mez-P{\'e}rez, A., Fernandez-Lopez, M.: The
  neon methodology framework: A scenario-based methodology for ontology
  development. Applied ontology  \textbf{10}(2),  107--145 (2015)

\bibitem{garcia_del_valle_dismanet_2021}
García~del Valle, E.P., Lagunes~García, G., Prieto~Santamaría, L., Zanin,
  M., Menasalvas~Ruiz, E., Rodríguez-González, A.: {DisMaNET}: {A}
  network-based tool to cross map disease vocabularies. Computer Methods and
  Programs in Biomedicine  \textbf{207},  106233 (Aug 2021).
  \doi{10.1016/j.cmpb.2021.106233}

\bibitem{vasant_ordo_2014}
Vasant, D., Chanas, L., Malone, J., Hanauer, M., Olry, A., Jupp, S., Robinson,
  P., Parkinson, H., Rath, A.: {ORDO}: {An} {Ontology} {Connecting} {Rare}
  {Disease}, {Epidemiology} and {Genetic} {Data}. In: Bio-Ontologies ISMB 2014
  (Jul 2014)

\bibitem{wishart_drugbank_2018}
Wishart, D.S., Feunang, Y.D., Guo, A.C., Lo, E.J., Marcu, A., Grant, J.R.,
  Sajed, T., Johnson, D., Li, C., Sayeeda, Z., Assempour, N., Iynkkaran, I.,
  Liu, Y., Maciejewski, A., Gale, N., Wilson, A., Chin, L., Cummings, R., Le,
  D., Pon, A., Knox, C., Wilson, M.: {DrugBank} 5.0: a major update to the
  {DrugBank} database for 2018. Nucleic Acids Research  \textbf{46}(D1),
  D1074--D1082 (Jan 2018). \doi{10.1093/nar/gkx1037}

\bibitem{zahn-zabal_nextprot_2020}
Zahn-Zabal, M., Michel, P.A., Gateau, A., Nikitin, F., Schaeffer, M., Audot,
  E., Gaudet, P., Duek, P.D., Teixeira, D., Rech de Laval, V., Samarasinghe,
  K., Bairoch, A., Lane, L.: The {neXtProt} knowledgebase in 2020: data, tools
  and usability improvements. Nucleic Acids Research  \textbf{48} (2020).
  \doi{10.1093/nar/gkz995}

\end{thebibliography}
